\documentclass[article,pdftex,10pt,a4paper,twocolumn]{article}
\usepackage[authoryear]{natbib}
\usepackage{graphics}

\usepackage{url}
\usepackage{epstopdf}
\usepackage{graphicx}
\usepackage{amsmath}
\usepackage{amssymb}

\usepackage{tabularx}
\usepackage{booktabs}

\usepackage{algorithm}
\usepackage{algpseudocode}
\usepackage{caption}
\usepackage{lineno}
\usepackage{microtype}

\usepackage[colorlinks=true,
            linkcolor=black,
            citecolor=black,
            urlcolor=black]{hyperref}

\RequirePackage{color}

\begin{document}

\title{CornViT: A Multi-Stage Convolutional Vision Transformer Framework for Hierarchical Corn Kernel Analysis}

\author{Sai Teja Erukude, Jane Mascarenhas and Lior Shamir \\ 
\small Kansas State University \\
\small 1701 Platt St, Manhattan, KS 66506, USA \\ 
\small erukude.saiteja@gmail.com, jane.mascarenhas.work@gmail.com, lshamir@mtu.edu}

\date{}
\maketitle

\begin{abstract}
Accurate grading of corn kernels is critical for seed certification, directional seeding, and breeding, yet it is still predominantly performed by manual inspection. This work introduces CornViT, a three-stage Convolutional Vision Transformer (CvT) framework that emulates the hierarchical reasoning of human seed analysts for single-kernel evaluation. Three sequential CvT-13 classifiers operate on $384 \times 384$ RGB images: Stage 1 distinguishes pure from impure kernels; Stage 2 categorizes pure kernels into flat and round morphologies; and Stage 3 determines the embryo orientation (up vs. down) for pure, flat kernels. Starting from a public corn seed image collection, we manually relabeled and filtered images to construct three stage-specific datasets: 7265 kernels for purity, 3859 pure kernels for morphology, and 1960 pure--flat kernels for embryo orientation, all released as benchmarks. Head-only fine-tuning of ImageNet-22k pretrained CvT-13 backbones yields test accuracies of 93.76\% for purity, 94.11\% for shape, and 91.12\% for embryo-orientation detection. Under identical training conditions, ResNet-50 reaches only 76.56 to 81.02 percent, whereas DenseNet-121 attains 86.56 to 89.38 percent accuracy. These results highlight the advantages of convolution-augmented self-attention for kernel analysis. To facilitate adoption, we deploy CornViT in a Flask-based web application that performs stage-wise inference and exposes interpretable outputs through a browser interface. Together, the CornViT framework, curated datasets, and web application provide a deployable solution for automated corn kernel quality assessment in seed quality workflows. Source code and data are publicly available.
\end{abstract}
\section{Introduction}

Maize (\emph{Zea mays
}) is a major global cereal crop used for food, feed, and bioenergy, and its kernel quality strongly influences germination, early vigor, and final yield. Kernel attributes such as physical purity, damage, and varietal consistency also affect downstream processes, including milling, grain fractionation, and industrial applications \cite{nehoshtan2021robust}. High standards of purity and kernel quality have been established by seed certification programs. However, the qualitative standards are still largely determined through a manual process that involves trained personnel inspecting, counting, and sorting kernels \cite{tia-0024-0013}. The~current processes of evaluating kernels are time-consuming, resource-intensive, and cannot be easily scaled to meet the demands of modern high-throughput breeding and commercial seed production. As a result, there is growing interest in automation, particularly using machine vision and deep learning techniques, to assist in automated seed evaluation and processing. That includes kernel counting, kernel defect detection, kernel breakage estimation, and seed vigor assessment.

Machine vision \cite{sonka2013image} and deep learning \cite{lecun2015deep} have rapidly advanced kernel-level inspection by providing scalable alternatives to manual evaluation. Early systems relied on handcrafted shape and texture descriptors paired with classical classifiers to distinguish whole from broken kernels or to assess purity. With modern deep learning, CNN-based \cite{o2015introduction} models have been developed for tasks such as on-ear kernel detection and counting, high-throughput phenotyping, and classification of good, defective, and impurity kernels \cite{9150684}. The 
 results demonstrated that deep feature extractors can match or surpass human graders on specific tasks. Self-supervised and contrastive learning approaches further reduce annotation costs by learning transferable kernel representations that support embryo-orientation detection and segmentation. At the same time, hyperspectral and RGB-based models have been used to predict maize variety, vigor, and germination \cite{10.3389/fpls.2023.1108355}. These studies collectively establish that image-based deep learning can capture many aspects of seed and kernel quality and form a strong foundation for automation.

Despite these advances, current image-based kernel analysis methods still face several major challenges. First, most rely on a single, monolithic classifier that maps directly from an image to a composite label such as ``good'', ``defective'', or ``impurity''. This~compresses diverse visual cues, purity, morphology, and orientation into one decision, making error analysis difficult, and provides limited interpretability for agronomists and seed analysts.Second, the field remains dominated by CNN-based and handcrafted feature pipelines that are effective for local pattern extraction but less suited to capturing the global shape context and subtle structural cues needed to distinguish borderline impurities or embryo orientation. Finally, many prior works address only one sub-task at a time (e.g., purity or embryo orientation) or are tightly coupled to specific imaging setups, which complicates reuse across quality dimensions or deployment environments. Although Vision Transformers (ViTs) \cite{dosovitskiy2020image} and related architectures offer improved capacity for modeling long-range relationships and have shown strong performance in plant disease detection and phenotyping \cite{Zhou2021MaizeIAS}, their use for fine-grained, kernel-level evaluation is still limited. In contrast, human graders naturally follow a structured, multi-stage reasoning process: they begin by assessing whether a kernel is pure, then determine its morphological category, and finally examine its orientation and finer anatomical features. Emulating this hierarchical decision pathway in automated models helps address both interpretability and modeling limitations in prior work.

At the same time, there is a broader trend toward deep learning-based image analysis in precision agriculture, which further motivates our approach. In crop-weed management, CWRepViT-Net has shown that encoder--decoder frameworks built from RepViT blocks can perform accurate semantic segmentation of crops and multiple weed species throughout the life cycle of soybean fields, highlighting the potential of transformer-style backbones for fine-grained canopy understanding and decision support in the field \cite{GOMROKI2025101472}. {In aerial monitoring, Succulent-YOLO combines a CLIP-enhanced YOLOv10 detector, dynamic group convolutions, and a multi-scale fusion neck with a Mamba-based super-resolution module (MambaIR) to detect succulent plants in UAV imagery, achieving high mean average precision even on low-resolution inputs} \cite{rs17132219}. {Closer to our application domain, Rocha et al. developed a real-time system that mounts a camera on a self-propelled forage harvester to capture images of chopped corn silage and uses machine learning to count whole kernels and estimate the Kernel Processing Score with strong agreement to laboratory sieve analysis, demonstrating that image-based models can provide actionable quality metrics directly in harvesting equipment} \cite{ROCHA2022107415}. Together, these examples show that deep learning, including transformer-based and CLIP-enhanced architectures, is increasingly used to automate image-based assessment of crops and grain products at canopy, field, and processing levels.

In this study, we investigate whether multi-stage Convolutional Vision Transformers (CvTs) \cite{wu2021cvt} can model the hierarchical decision-making process used by human seed analysts for single-kernel evaluation. To this end, we introduce CornViT, a three-step CvT-based framework that sequentially classifies kernel purity, morphological type, and embryo orientation. Each stage employs an independently fine-tuned CvT-13 \mbox{model \cite{microsoft_CvT_github}} operating on $384 \times 384$ RGB images, leveraging CvT’s combination of convolutional inductive biases with the global context modeling of Transformers. By decomposing grading into three explicit decisions and using a transformer-based backbone, CornViT is designed to address the limitations of prior work: it replaces monolithic labels with interpretable intermediate outputs, uses convolution-augmented self-attention to capture both local surface cues and global kernel shape, and provides modular stages that can be adapted to different imaging setups or extended to additional quality attributes. Across the three stages, CornViT substantially outperforms strong CNN baselines: ResNet-50 \cite{he2016deep, koonce2021resnet} attains only around 77 to 81 percent accuracy and DenseNet-121 \cite{huang2017densely} around {87 to 89 percent}, whereas CornViT achieves 93.76\%, 94.11\%, and 91.12\% test accuracy for purity, shape, and embryo-orientation classification, respectively. This~highlights the advantages of transformer-based architectures for fine-grained agricultural vision tasks.  We deploy the models in a lightweight web application that supports stage-wise inference and exposes interpretable outputs through a simple browser interface, illustrating how such a framework can be integrated into seed-quality and precision-agriculture workflows.

Three major contributions from this research are as follows:
\begin{enumerate}
    \item Introduction of CornViT, a three-stage CvT-based framework that mirrors human-style hierarchical reasoning for kernel purity, morphology, and embryo orientation.
    \item Construction and release of a stage-wise annotated corn kernel dataset comprising three curated subsets aligned with the purity, shape, and embryo-orientation tasks.
    \item Development of a ready-to-use web application that exposes the full CornViT pipeline through a browser interface, enabling easy adoption in seed quality assessment and precision-agriculture workflows.
\end{enumerate}

The remainder of this paper describes related work (Section~\ref{sec_related_work}), materials and methods (Section~\ref{sec_methods}), experimental results (Section~\ref{sec_results}), the web application integration (Section~\ref{sec_web_app}), a detailed discussion (Section~\ref{sec_discussion}), and conclusions (Section~\ref{sec_conclusions}).

\section{Background and Related Work}
\label{sec_related_work}

\subsection{Automated Inspection of Maize Kernels and Seeds}

Within this space, several studies have focused specifically on corn and other crop seeds. Liao et al.\ developed an early machine vision system for automated corn quality inspection in which binary images of kernels were converted into one-dimensional shape profiles; key shape parameters were then extracted and fed into a neural network. Their model accurately distinguished whole from broken kernels, achieving up to 99\% accuracy for whole kernels and up to 96\% for broken kernels \cite{Liao1993CornKernelBreakage}. More recently, Qingzhen \mbox{Zhu et al}.\ presented a hyperspectral imaging-based maize variety identification approach that combines SG and SNV preprocessing, CARS wavelength selection, and a CNN–LSTM model, achieving 95.27\% accuracy and outperforming traditional chemometric methods, demonstrating its effectiveness for efficient and reliable maize variety classification \cite{Zhu2025MaizeSeedVariety}.

Related ideas have also been applied to other crops. For~example, Prashant Yawalkar and colleagues developed an automated onion seed quality assessment system using image processing and machine learning, showing that CNN-based deep learning outperforms traditional methods in both accuracy and scalability, and providing a faster, more reliable alternative to manual inspection for precision agriculture \cite{surse2025automated}.

More recent work has explored modern representation learning for maize kernel tasks. David Dong demonstrated that self-supervised contrastive learning methods (NNCLR and SimCLR) \cite{chen2020simple} enable highly efficient maize kernel embryo-orientation classification and segmentation, outperforming supervised baselines, transferring effectively across tasks, and achieving strong accuracy (DICE 0.81) even with as little as 1\% annotated data \cite{10.3389/fpls.2023.1108355}. These results indicate that kernel-level inspection tasks are well-suited to contemporary deep and representation-learning pipelines.

\subsection{Embryo Orientation and Directional Seeding}

Seed orientation plays a significant role in optimizing corn growth and yield, as several studies have shown that directing seeds during planting can influence leaf azimuth, canopy structure, and ultimately light interception. Peters and Woolley (1959) first noted that seed direction affects leaf orientation \cite{peters1959orientation}, and Fortin and Pierce (1996) demonstrated that controlled seed orientation can align a majority of ear leaves with embryo direction \cite{fortin1996leaf}. Field trials by Toler et al.\ (1999) further showed that manipulating seed orientation to achieve across-row leaf alignment increased yields by {10 to 20 percent} by improving light interception and reducing competition \cite{toler1999corn}. Similarly, Torres et al.\ (2011
) reported that specific seed placements, such as tips down or embryos up, promoted favorable leaf orientation and significantly enhanced light interception and yield \cite{torres2011maize}.

Motivated by these agronomic benefits, several imaging-based systems have been proposed to automatically determine and adjust embryo orientation. Yingbiao Wang introduced a color-based image processing and contour-analysis method that automatically identifies maize seed embryo orientation with over 95\% accuracy and an average error of 2.2$^\circ$ across three maize varieties \cite{10.1007/978-3-642-54341-8_1}. Huijuan Bai et al.\ developed an identification and adjustment prototype that orients corn seeds with the embryo side upward and the long axis perpendicular to the row, using image processing and pneumatic adjustment to achieve an 84\% success rate in seed orientation \cite{bai2024image}. Liu Changqing et al.\ proposed another image-based method that detects the embryo by its whitish color and position along the major axis, then determines tip direction to calculate the orientation angle, achieving 97.1\% \mbox{accuracy \cite{changqing2015dynamic}}. These systems typically rely on handcrafted geometric and color-based features within device-specific imaging setups, which can make them sensitive to variations in lighting, seed appearance, and camera conditions.

More recently, David Dong’s work on self-supervised contrastive learning showed that representation-learning approaches can further improve data efficiency and robustness for embryo-orientation classification and segmentation \cite{10.3389/fpls.2023.1108355}. In contrast to handcrafted pipelines, our CornViT framework adopts a multi-stage Vision Transformer architecture that autonomously learns hierarchical features, explicitly mirrors the purity--shape--orientation decision sequence used by human graders, and aims to provide a more robust, accurate, and scalable system for modern seed phenotyping and directional seeding applications.

\subsection{Vision Transformers in Precision Agriculture}

Transformers are increasingly being adopted in agriculture and plant phenotyping, with many recent studies showing that Vision Transformers (ViTs) and their variants often match or surpass CNNs in tasks such as disease detection, weed identification, and \mbox{crop monitoring}.

In the GreenViT framework, ViTs were used for plant disease detection by dividing images into sequential patches, achieving superior performance compared to state-of-the-art CNN models on benchmark datasets \cite{parez2023visual}. A ViT-based smartphone application, ViT-SmartAgri, was developed to classify 10 tomato disease classes from 10,010 leaf images, achieving 90.99\% accuracy and outperforming Inception V3, demonstrating its potential for large-scale smart agriculture \cite{barman2024vit}. Guoqiang Li et al.\ employed a lightweight MobileViT architecture enhanced with inverted residual blocks and a CBAM attention module to efficiently capture long-range dependencies and focus on relevant features, enabling accurate, real-time plant disease detection on mobile devices \cite{li2023pmvt}. A ViT-B16 backbone has also been applied to multispectral plant images for disease identification and classification, achieving the highest performance among tested models and highlighting its potential for improving crop disease management \cite{de2023multispectral}. Likewise, a ViT model applied to the DeepWeeds dataset achieved 96.9\% accuracy across nine weed species, demonstrating the effectiveness of transformer-based architectures for weed detection in agricultural settings \cite{hasasneh2025weednet}.

A recent review by Zhang et al.\ highlights a broader shift from CNN-only approaches toward ViT and hybrid CNN-transformer models for pest and disease detection, emphasizing that transformer-based architectures offer enhanced global reasoning and feature representation \cite{zhang2025convolutional}. Despite these advances, the use of ViTs in seed and kernel quality assessment remains limited, as most studies still depend on CNN-based approaches. This~emerging evidence from canopy- and leaf-level tasks nonetheless motivates exploring ViT-style models for fine-grained, kernel-level analysis, such as in the present work.

\subsection{Multi-Stage and Hierarchical Deep Learning Pipelines}

The concept of multi-stage classification pipelines has been applied in several domains where complex decisions are naturally hierarchical. For~example, in cancer grading, a study on Gleason system grading employed a multi-stage deep learning pipeline with multilevel binary CNN classifiers to progressively identify Gleason patterns, scores, and grade groups from digitized prostate biopsy images, demonstrating the effectiveness of staged classification in complex image analysis tasks \cite{hammouda2022multi}. In a different context, short-term wind power forecasting has been tackled with a multi-stage, hierarchical, deep learning pipeline that uses layered information-fusion modules to progressively integrate homogeneous and heterogeneous SCADA data, improving the model’s ability to capture complex temporal and spatial relationships and enhancing prediction accuracy \cite{qin2025hierarchical}.

More broadly, hierarchical and cascade classifiers have long been used in machine vision applications that require separating easy from difficult examples, identifying subcategories based on specific object characteristics (e.g., morphology), or implementing coarse-to-fine recognition strategies. However, such structured model hierarchies are still rare in seed and kernel inspection, even though expert humans naturally use hierarchical decision processes in this field, for example, first determining whether a sample is pure or impure, then assessing its morphology, and finally examining its orientation and finer anatomical cues.

CornViT extends these ideas to corn kernel grading with a three-stage, task-specific design that explicitly mirrors this human reasoning sequence. Each stage is implemented with a CvT-based architecture, allowing the framework to leverage convolution-augmented self-attention for both local texture analysis and global shape reasoning.

\section{Materials and Methods} 


\label{sec_methods}

\subsection{Problem Formulation}

This study considers single-kernel RGB images in which each image contains exactly one corn kernel on a uniform background. The~goal is to obtain a hierarchical description of the kernel through three binary decisions:

\begin{enumerate}
    \item Purity (Stage 1):
    \begin{itemize}
        \item Pure: visually acceptable kernels without obvious defects or discoloration.
        \item Impure: kernels that are broken, discolored, silkcut {(intact kernels with visible surface cracks and silk-embedded fissures)}, or otherwise unsuitable.
    \end{itemize}

    \item Shape (Stage 2, conditioned on purity):
    \begin{itemize}
        \item Flat: kernels with a flattened dorsal--ventral profile.
        \item Round: kernels with a more equiaxed or rounded morphology.
    \end{itemize}

    \item Embryo orientation (Stage 3, conditioned on purity and flat shape):
    \begin{itemize}
        \item Embryo up: the embryo side faces the camera.
        \item Embryo down: the embryo faces away from the camera.
    \end{itemize}
\end{enumerate}

{Let} 
 $\mathbf{x} \in \mathbb{R}^{H \times W \times 3}$ denote an RGB kernel image, and let $f_1, f_2, f_3$ denote the Stage 1--3 models. The~overall pipeline maps
\[
\mathbf{x} \;\mapsto\; (y_1, y_2, y_3),
\]
where $y_1 \in \{\text{pure}, \text{impure}\}$, 
$y_2 \in \{\text{flat}, \text{round}\}$ (if defined), 
$y_3 \in \{\text{embryo up}, \text{embryo down}\}$ (if defined).

For kernels that are predicted as impure in Stage 1, no further classification is attempted, so $y_2$ and $y_3$ remain undefined. Similarly, for kernels that are pure but round, Stage 3 is skipped and $y_3$ is undefined.

In practice, this hierarchy is implemented using three independent binary classifiers, each trained on a stage-specific dataset: $D_1$ for purity, $D_2$ for shape (pure kernels only), and $D_3$ for embryo orientation (pure--flat kernels only). This~design allows each stage to specialize in its own decision boundary while maintaining a simple and interpretable global pipeline.

\begin{figure*}[ht]
    \centering
    \includegraphics[width=5in]{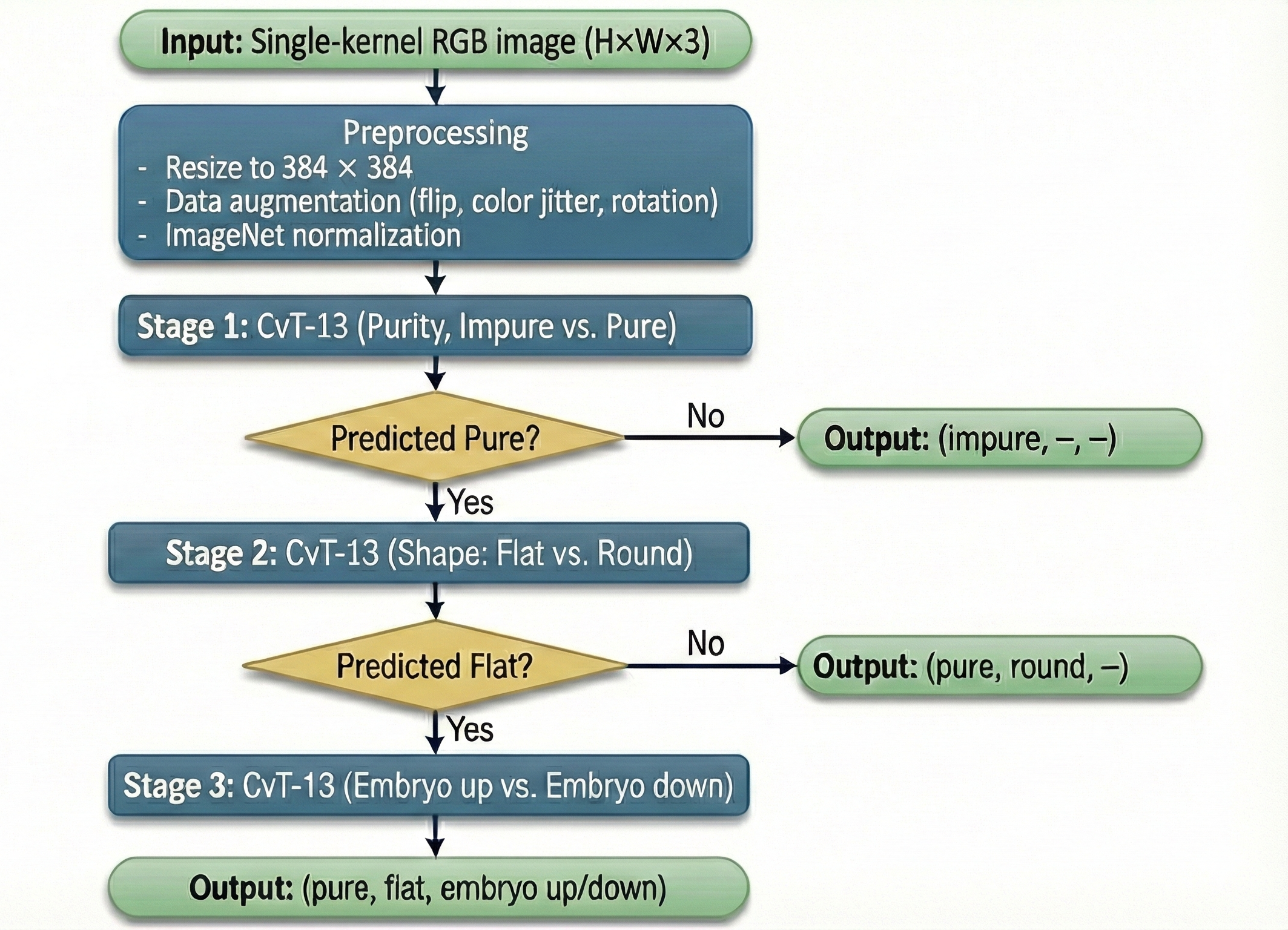}
    \caption{The 
 overall workflow of the proposed CornViT system illustrates how a single-kernel RGB image passes through the preprocessing module and then sequentially through three \mbox{CvT-13 classifiers}.}
    \label{fig_flow_chart}
\end{figure*}

The overall workflow of the proposed CornViT framework is summarized in the flowchart in Figure \ref{fig_flow_chart}. Starting from a single-kernel RGB image on a uniform background, the image is first passed through a standardized preprocessing pipeline (resize, augmentation, and ImageNet-style normalization). The~preprocessed image is then processed sequentially by three CvT-13 classifiers corresponding to Stage 1 (purity), Stage 2 (shape), and Stage 3 (embryo orientation). At each stage, a binary decision is made, and the hierarchy either terminates (for impure or pure-round kernels) or progresses to the next classifier (for pure and pure–flat kernels). The~final output is a hierarchical label tuple ($y_1$, $y_2$, $y_3$) that describes purity, morphology, and embryo orientation, with undefined components for skipped stages.

\subsection{Dataset Preparation}

A publicly available corn seed image dataset \cite{sethu123123_cornseed_2025} hosted on Kaggle served as the starting point for this study. The~original dataset contains four labeled classes: broken, discolored, pure, and silkcut, but closer inspection revealed substantial inconsistencies between the class labels and corresponding images. Several images were misplaced across folders and did not correctly represent their annotated class.  

Moreover, the predefined class structure did not align with the hierarchical purity, shape, and embryo-orientation objectives considered here. Due to these inconsistencies, the dataset was unsuitable for direct model training. Furthermore, publicly available image datasets for corn kernel analysis are extremely limited, particularly for classification tasks involved in this study.

To address this gap, we performed a comprehensive manual curation process in which each image was visually inspected, hand-picked, and reassigned to its correct class. In total, 17,801 single-kernel images from the Kaggle download were examined. Of these, \mbox{10,536 images} were duplicates and were therefore discarded. A total of 7265 images were retained in the curated pool used to construct the stage-wise datasets (Tables \ref{tab_dataset_stage1}--\ref{tab_dataset_stage3}). Approximately 50\% of these retained images required relabeling their class during curation.

Using this cleaned pool of images, we then constructed three progressively filtered datasets, each tailored to a specific stage of the CornViT pipeline. We regard these three curated datasets as a key contribution of this work. They provide a ready-to-use benchmark suite for researchers interested in corn kernel purity, morphology, and embryo orientation. The~full datasets, along with train/validation/test splits, will be made publicly available at \url{https://doi.org/10.5281/zenodo.17693853}.

\subsubsection {Stage 1 Dataset: Purity Classification}

\begin{figure}[ht]
    \centering
    \includegraphics[width=3in]{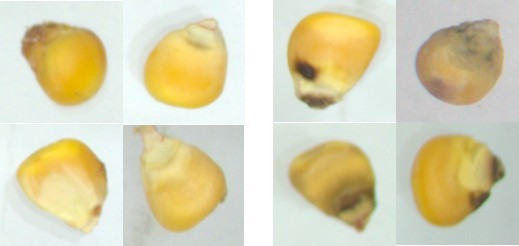}
    \caption{A sample of images randomly picked from the Stage 1 dataset, showing pure kernels on the left and impure kernels on the right.}
    \label{fig_dataset_stage1}
\end{figure}

The first stage aims to distinguish pure kernels from impure ones. To this end, the curated dataset was reorganized into two classes: (1) Pure: kernels that are visually free from defects or discoloration, (2) Impure: an aggregated class combining broken, discolored, and silkcut kernels. Figure~\ref{fig_dataset_stage1} presents sample images, and Table~\ref{tab_dataset_stage1} summarizes the train/validation/test partitioning, which follows a 70/15/15 split.

\begin{table}[ht] 
    \small
    \caption{Summary 
 of the Stage 1 dataset, showing the number of pure and impure samples in the training, validation, and test subsets.}
    \label{tab_dataset_stage1}
    \begin{tabularx}{\columnwidth}{cccc}
        \toprule
        \textbf{Subset}	& \textbf{Pure}	& \textbf{Impure}  	& \textbf{Total}\\
        \midrule
        Training Set    & 2586                & 2499                     & 5085      \\ 
        Validation Set  & 555                  & 535                       & 1090      \\ 
        Test Set        & 554                  & 536                       & 1090      \\
      \midrule
        Overall Total 
 & 3695         & 3570                     & 7265      \\
        \bottomrule
    \end{tabularx}
\end{table}

\subsubsection{Stage 2 Dataset: Morphological Classification}

\begin{figure}[ht]
    \centering   
    \includegraphics[width=3in]{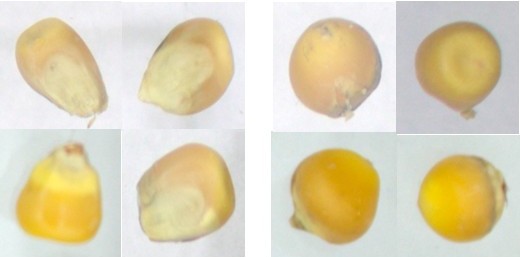}
    \caption{Representative samples from the Stage 2 dataset showing flat kernels on the left and round kernels on the right.}
    \label{fig_dataset_stage2}
\end{figure}

The second stage focuses on morphological categorization of kernels based on shape characteristics. Since impure kernels are not relevant for further morphological or orientation analysis, only the pure samples from Stage 1 were included in Stage 2. From these, two new classes were identified: (1) Flat and (2) Round. This~refinement is motivated by the observation that embryo orientation, the target of Stage 3, is only visually meaningful for flat kernels. Figure~\ref{fig_dataset_stage2} presents sample images, and Table~\ref{tab_dataset_stage2} summarizes the \mbox{train/validation/test split}.

\begin{table}[ht] 
    \small
    \caption{Summary of the Stage 2 dataset, showing the number of flat and round samples in the training, validation, and test subsets following the 70/15/15 split.}
    \label{tab_dataset_stage2}
    \begin{tabularx}{\columnwidth}{cccc}
        \toprule
        \textbf{Subset}	& \textbf{Flat}	& \textbf{Round}  	& \textbf{Total}\\
        \midrule
        Training Set    & 1374                & 1329                     & 2703      \\ 
        Validation Set  & 294                  & 284                       & 578      \\ 
        Test Set        & 294                  & 284                       & 578      \\
      \midrule
        Overall Total   & 1962         & 1897                     & 3859      \\
        \bottomrule
    \end{tabularx}
\end{table}

\subsubsection{Stage 3 Dataset: Embryo Orientation Classification}

\begin{figure}[ht]
    \centering
    \includegraphics[width=3in]{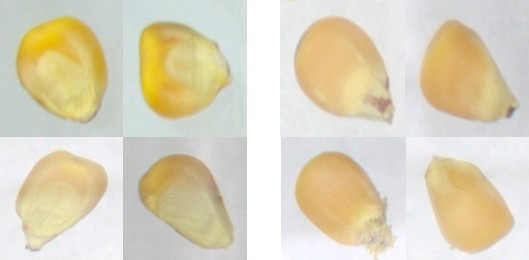}
    \caption{Sample images randomly picked from the Stage 3 dataset depicting embryo orientation, with kernels exhibiting the embryo-up class on the left and the embryo-down class on the right.}
    \label{fig_dataset_stage3}
\end{figure}

The final stage addresses embryo orientation detection, which is critical for kernel viability and downstream seed processing. Because embryo orientation is visible only on flat kernels, Stage 3 uses the flat subset of the Stage 2 dataset as its base. Two classes were manually derived: (1) Embryo Up and (2) Embryo Down. Figure~\ref{fig_dataset_stage3} presents sample images, and Table~\ref{tab_dataset_stage3} summarizes the train/validation/test split.

\begin{table}[ht] 
    \small
    \caption{Summary of the Stage 3 dataset, showing the number of embryo-up and embryo-down samples in the training, validation, and test subsets following the 70/15/15 split.}
    \label{tab_dataset_stage3}
    \begin{tabularx}{\columnwidth}{Xccc}
        \toprule
        \textbf{Subset}	& \textbf{Embryo Up}	& \textbf{Embryo Down}  	& \textbf{Total}\\
        \midrule
        Training    & 813                  & 561                     & 1374      \\ 
        Validation  & 174                  & 119                       & 293      \\ 
        Test        & 174                  & 119                       & 293      \\
       \midrule
        Total   & 1161         & 799                    & 1960      \\
        \bottomrule
    \end{tabularx}
\end{table}

\subsection{Image Preprocessing}
\label{sec_image_preprocessing}

All images were processed through a standardized preprocessing pipeline before being fed into the CornViT models. Each image was first resized to a fixed resolution of $384 \times 384$ to match the CvT-13 backbone configuration.

To improve model robustness and mitigate overfitting, we applied a set of common on-the-fly data augmentations during training. These augmentations included random horizontal and vertical flips, color jittering (adjustments to brightness, contrast, and saturation), 
and small in-plane rotations of up to $15^\circ$. After augmentation, each image was converted into a normalized PyTorch tensor. Normalization followed the standard ImageNet statistics (mean = $[0.485,\, 0.456,\, 0.406]$, standard deviation = $[0.229,\, 0.224,\, 0.225]$), which is typically used for models pretrained on ImageNet. 

Overall, this preprocessing and augmentation pipeline ensures consistent image scaling, improves generalization under modest domain shifts, and provides well-conditioned inputs for all three CornViT stages. The~full implementation is available in the accompanying code repository 
 \url{https://github.com/SaiTeja-Erukude/CornViT} (accessed on 19 December 2025).

\subsection{CornViT Architecture}

Each stage of the CornViT framework is implemented as an independent Convolutional Vision Transformer (CvT-13) classifier \cite{wu2021cvt} built on top of the official Microsoft CvT implementation. All~three stages share the same backbone architecture but are trained with their own binary classification heads and datasets.

\subsubsection{Convolutional Vision Transformer (CvT-13)}

The Convolutional Vision Transformer (CvT) \cite{wu2021cvt} is a hybrid vision backbone that combines the global self-attention of Vision Transformers (ViTs) \cite{dosovitskiy2020image} with the local inductive biases of CNNs. CvT introduces two key modifications to the vanilla ViT architecture: \mbox{(1) Convolutional} token embedding and (2) Convolutional projection in self-attention.

Instead of partitioning the image into non-overlapping patches and flattening them with a linear projection (as in ViT), CvT uses convolutional layers to generate tokens. These convolutions define local receptive fields and perform spatial down-sampling, allowing each stage to operate on progressively coarser yet semantically richer feature maps. This~injects CNN-like properties such as shift, scale, and distortion invariance into \mbox{the transformer}.

In the transformer blocks, CvT replaces pure linear projections for queries, keys, and values with convolutional projections. This enables the attention mechanism to be aware of local spatial neighborhoods while still modeling long-range dependencies through multi-head self-attention. As a result, CvT can better capture fine-grained textures (e.g., kernel surface cues) and global shape simultaneously, often with fewer parameters and FLOPs than comparable ViT or deep CNN backbones.

\begin{figure*}[h]
    \centering
    \includegraphics[width=6in]{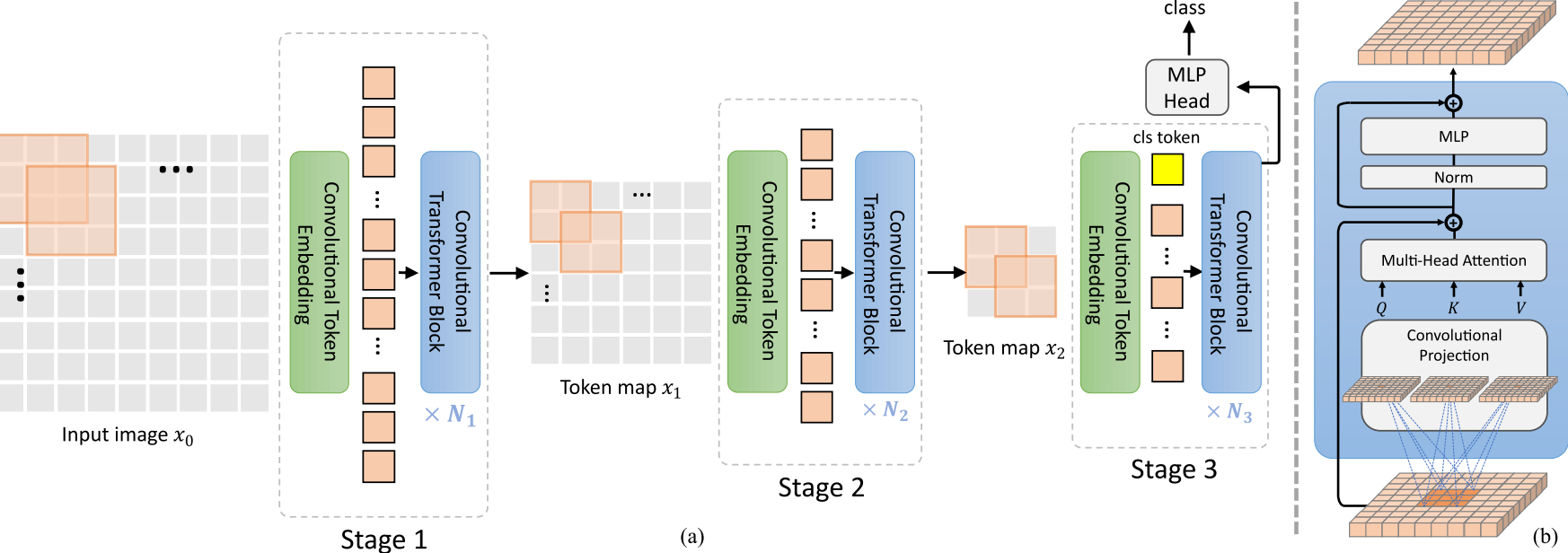}
    \caption{Schematic 
 of the CvT-13 architecture used as the backbone in all CornViT stages \cite{wu2021cvt}. (a) The complete architecture highlights the hierarchical multi-stage design facilitated by the Convolutional Token Embedding layer. (b) A detailed illustration of the Convolutional Transformer Block, which begins with a convolutional projection layer.}
    \label{fig_cvt_pipeline}
\end{figure*}

CvT-13 is the smallest variant described by Wu et al., with three transformer stages of increasing channel width and decreasing spatial resolution as depicted in Figure~\ref{fig_cvt_pipeline}. For~$384 \times 384$ inputs, the final stage produces a compact global representation that feeds a lightweight classification head. This~makes CvT-13 a good fit for our single-kernel classification setting, where both local surface detail and global kernel morphology \mbox{are important}.

\begin{figure}[ht]
    \includegraphics[width=3in]{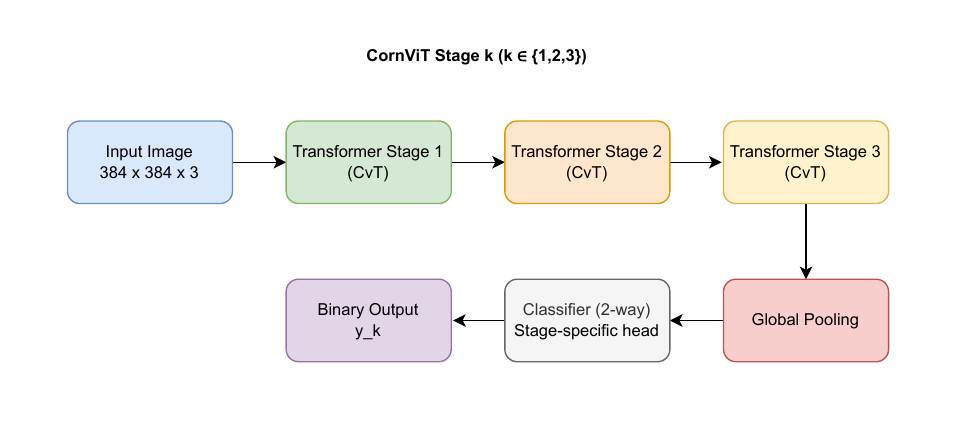}
    \caption{Block diagram of the proposed CornViT network showing internal structure of a \mbox{single stage}.}
    \label{fig_block_diagram}
\end{figure}

A block-level overview of the proposed CornViT network is shown in Figure \ref{fig_block_diagram}, which details the internal structure of a single stage: three transform stages of the CvT backbone, followed by a global pooling module and a 2-unit stage-specific classification head. All~stages of CornViT follow the same internal structure.

\subsubsection{Experimental Setup}

All stages in the proposed framework are implemented using the official Microsoft CvT codebase \cite{microsoft_CvT_github} {with the CvT-13 configuration for 384 $\times$ 384 inputs. For~each stage, we initialize a CvT-13 backbone from the publicly available ImageNet-22k pretrained checkpoint} \cite{cvt13_imagenet22k_checkpoint} {and adapt it to the corresponding binary task by attaching a 2-unit classification head. Low-level configuration details (e.g., repository cloning and config files) follow the official implementation and are documented in the public code repository.}

{We adopt a head-only fine-tuning strategy in which the CvT backbone remains frozen and only the final linear classification layer is updated. This~choice was motivated by three factors: (i) the curated stage-specific datasets are relatively small compared with large-scale vision corpora, increasing the risk of overfitting when unfreezing deeper transformer blocks; (ii) head-only tuning substantially reduces training time and GPU memory requirements, enabling fully independent training for the three stages; (iii) freezing the backbone ensures stable, comparable feature representations across purity, morphology, and embryo-orientation tasks. Although partial unfreezing (e.g., unfreezing the final transformer stage) may offer additional performance gains—particularly for the visually subtle Stage 3 embryo-orientation task---we leave this investigation for future work.}

{All stages share the same training configuration, including optimizer, learning-rate scheduler, label smoothing, and number of epochs; the full set of hyperparameters is summarized in Table}~\ref{tab_training_config}. The~entire training and inference pipeline is implemented in PyTorch (version 2.9.0) 
 \cite{paszke2019pytorch}, and the complete source code is publicly available at: 
 \url{https://github.com/SaiTeja-Erukude/CornViT} (accessed on 19 December 2025).

\begin{table*}[ht]
\centering
\small
\caption{Summary of training configuration for all CornViT stages.}
\label{tab_training_config}
    \begin{tabularx}{0.82\textwidth}{cc}
    \toprule
    \textbf{Component} & \textbf{Setting} \\
    \midrule
        Backbone & CvT-13 (official Microsoft implementation) \cite{microsoft_CvT_github} \\
        Input resolution & \(384 \times 384\) \\
        Number of classes & 2 (binary classification) \\
        Pretrained weights & ImageNet-22k checkpoint (OneDrive link) \cite{cvt13_imagenet22k_checkpoint} \\
        Fine-tuning strategy & Head-only (backbone frozen) \\
        Optimizer & AdamW \cite{loshchilov2017decoupled} \\
        Learning rate & \(1 \times 10^{-4}\) \\
        Weight decay & 0.05 \\
        Loss function & SoftTargetCrossEntropy \\
        Label smoothing & 0.1 \\
        Learning-rate scheduler & CosineLRScheduler \\
        Warm-up & 5 epochs, warmup\_lr\_init = \(1 \times 10^{-5}\) \\
        Minimum LR & \(1 \times 10^{-6}\) \\
        Total epochs & 20 \\
        Implementation & PyTorch \cite{paszke2019pytorch} \\
        Repository & \url{https://github.com/SaiTeja-Erukude/CornViT} \\
    \bottomrule
    \end{tabularx}
\end{table*}

\subsection{Algorithms}

The full training and inference procedures of the proposed CornViT framework are summarized using two stage-wise algorithms. Algorithm~\ref{alg_cornvit_training} describes the independent training strategy for each classification stage, and Algorithm~\ref{alg_cornvit_inference} outlines the hierarchical inference pipeline used to obtain the final kernel labels.

{For clarity, we recall that $D_1$, $D_2$, and $D_3$ denote the three curated datasets used in this work: $D_1$ contains all kernels used for purity classification; $D_2$ is a subset of $D_1$ that contains only kernels labeled as pure and is used for shape classification; and $D_3$ is a subset of $D_2$ that contains only pure–flat kernels and is used for embryo-orientation classification.}

\begin{algorithm*}[ht]
\caption{CornViT 
 Stage-wise Training}
\label{alg_cornvit_training}
\begin{algorithmic}[1]
    \Require \\
    $D_1 = \{(x_i, y^1_i)\}$: Stage 1 data {(impure vs. pure)} \\
    $D_2 = \{(x_j, y^2_j)\}$: Stage 2 data {(flat vs. round; pure only)} \\
    $D_3 = \{(x_k, y^3_k)\}$: Stage 3 data {(embryo up vs. down; pure--flat only)} \\
    CvT-13 models $f_1, f_2, f_3$ with 2-class heads \\
    \texttt{train\_transforms}, \texttt{val\_transforms}, number of epochs $T = 20$
\For{$s \in \{1,2,3\}$} \Comment{stage-wise training loop}
    \State Initialize $f_s$ with ImageNet-pretrained CvT-13 weights
    \State Replace final classification head with a 2-unit linear layer
    \State Freeze all backbone parameters of $f_s$ (head-only fine-tuning)

    \If{$s = 1$}
        \State $D_s \gets D_1$
    \ElsIf{$s = 2$}
        \State $D_s \gets D_2$
    \Else
        \State $D_s \gets D_3$
    \EndIf

    \For{$\text{epoch} = 1$ \textbf{to} $T$}
        \ForAll{$(x, y)$ in training split of $D_s$}
            \State $x_{\text{aug}} \gets \texttt{train\_transforms}(x)$
            \State $p \gets f_s(x_{\text{aug}})$
            \State $L \gets \texttt{SoftTargetCrossEntropy}(p, y)$
            \State Update head parameters of $f_s$ using AdamW to minimize $L$
        \EndFor

        \State Evaluate $f_s$ on the validation split of $D_s$ using \texttt{val\_transforms}
    \EndFor
\EndFor
\end{algorithmic}
\end{algorithm*}

\begin{algorithm}[ht]
\caption{CornViT 
 Hierarchical Inference}
\label{alg_cornvit_inference}
\begin{algorithmic}[1]
    \Require \\
    $x_{\text{test}}$: RGB image of a single kernel \\
    \hspace*{1.65em} Load trained models $f_1, f_2, f_3$ \\
    \hspace*{1.65em} \texttt{val\_transforms}
    
    \State $x_1 \gets \texttt{val\_transforms}(x_{\text{test}})$
    
    \State \textbf{Stage 1}: Pure vs. Impure
    \State $\hat{y}_1 \gets f_1(x_1)$
    \If{$\hat{y}_1 = \text{impure}$}
        \State \textbf{Output:} (impure, --, --)
        \State \Return
    \EndIf
    
    \State \textbf{Stage 2}: Flat vs. Round
    \State $x_2 \gets x_1$
    \State $\hat{y}_2 \gets f_2(x_2)$
    \If{$\hat{y}_2 = \text{round}$}
        \State \textbf{Output:} (pure, round, --)
        \State \Return
    \EndIf
    
    \State \textbf{Stage 3}: Embryo Up vs. Down
    \State $x_3 \gets x_2$
    \State $\hat{y}_3 \gets f_3(x_3)$
    \State \textbf{Output:} (pure, flat, $\hat{y}_3$)
\end{algorithmic}
\end{algorithm}

\subsection{Evaluation Metrics}

To comprehensively assess the performance of each stage in the proposed CornViT framework, we employ standard classification metrics, including Accuracy (Acc), Precision (Pre), Recall (Re), F1 score, and their macro and weighted averages. Since each stage is formulated as a binary classification problem, these metrics characterize different aspects of the model's behavior, such as overall correctness, reliability on the positive class, and robustness under class imbalance \cite{juba2019precision}.

Let $\mathrm{TP}$, $\mathrm{FP}$, $\mathrm{TN}$, and $\mathrm{FN}$ denote the number of true positives, false positives, true negatives, and false negatives, respectively. 

The Accuracy 
measures the overall proportion of correctly classified samples.
\begin{linenomath}
\begin{equation}
    \mathrm{Acc} = \frac{\mathrm{TP} + \mathrm{TN}}{\mathrm{TP} + \mathrm{TN} + \mathrm{FP} + \mathrm{FN}}
\end{equation}
\end{linenomath}

Precision measures the proportion of predicted positive samples that are actually positive.
\begin{linenomath}
\begin{equation}
    \mathrm{Pre} = \frac{\mathrm{TP}}{\mathrm{TP} + \mathrm{FP}}
\end{equation}
\end{linenomath}

Recall (also known as sensitivity) quantifies the proportion of actual positive samples that are correctly identified.
\begin{linenomath}
\begin{equation}
    \mathrm{Re} = \frac{\mathrm{TP}}{\mathrm{TP} + \mathrm{FN}}
\end{equation}
\end{linenomath}

The F1 score is the harmonic mean of Precision and Recall.
\begin{linenomath}
\begin{equation}
    \mathrm{F1} = 2 \cdot \frac{\mathrm{Pre} \cdot \mathrm{Re}}{\mathrm{Pre} + \mathrm{Re}}
\end{equation}
\end{linenomath}

For completeness, we also report macro and weighted averages across classes. 
Let $C$ be the number of classes (here, $C=2$), and let $\mathrm{Pre}_c$, $\mathrm{Re}_c$, and $\mathrm{F1}_c$ denote the class-wise metrics for class $c$, with $n_c$ samples in class $c$ and $N = \sum_{c=1}^{C} n_c$ the total number of samples.

The macro-average of a generic metric $M \in \{\mathrm{Pre}, \mathrm{Re}, \mathrm{F1}\}$ is computed as
\begin{linenomath}
\begin{equation}
    M_{\mathrm{macro}} = \frac{1}{C} \sum_{c=1}^{C} M_c,
\end{equation}
\end{linenomath}
which treats all classes equally. 

The weighted-average version accounts for class imbalance by weighting each class by its support:
\begin{linenomath}
\begin{equation}
    M_{\mathrm{weighted}} = \sum_{c=1}^{C} \frac{n_c}{N} \, M_c.
\end{equation}
\end{linenomath}

In our experiments, we report Accuracy together with macro- and weighted-average
Precision, Recall, and F1, providing a balanced view of performance under potentially imbalanced class distributions at each CornViT stage. {Model selection for each stage is based primarily on validation Accuracy, while also inspecting macro- and weighted-average F1; when candidate models achieve similar validation Accuracy, we prefer those with higher macro-F1 to avoid degrading minority-class performance.}

\section{Results}
\label{sec_results}

\subsection{Baseline CNNs}

To establish performance benchmarks and contextualize the performance of the proposed CornViT framework, baseline experiments were carried out. Two convolutional neural networks (CNNs) \cite{o2015introduction} were utilized: ResNet-50 \cite{he2016deep, koonce2021resnet} and DenseNet-121 \cite{huang2017densely}. Both models were chosen because they represent strong, well-established backbones for image classification and have been extensively used in agricultural and plant-phenotyping tasks.

{A broader survey including additional lightweight CNNs such as EfficientNet or MobileNet
would be valuable, but lies outside the scope of this study. Our goal in this work is not to exhaustively benchmark all CNN variants, but rather to compare a representative pair of strong, widely used CNN backbones against the proposed multi-stage CvT framework on a newly curated kernel dataset. Exploring a broader range of mobile-oriented CNNs is therefore left as complementary future work, orthogonal to the main question of whether CvT offers advantages for hierarchical kernel-level analysis.}

ResNet-50 is a 50-layer residual network that introduces identity-based skip connections (residual blocks) to ease the optimization of deep models and mitigate vanishing-gradient issues. Each residual block learns a residual mapping with respect to its input, allowing gradients to flow more directly through the network and enabling effective training of very deep architectures \cite{koonce2021resnet}.

DenseNet-121 is a densely connected convolutional network in which each layer receives, as input, the concatenation of all feature maps from preceding layers within the same dense block. This~design encourages feature reuse, improves information flow, and reduces the number of parameters compared to traditional feed-forward CNNs with comparable depth \cite{huang2017densely}.

For all three CornViT stages (purity, shape, embryo orientation), the baselines were configured as follows:

\begin{itemize}
    \item Both ResNet-50 and DenseNet-121 were initialized from ImageNet-pretrained weights.
    \item The final classification layers were replaced with new 2-class fully connected heads, matching the binary tasks at each stage.
    \item The same preprocessing and data augmentation pipeline described in Section~\ref{sec_image_preprocessing} was applied (resize, random flips, color jitter, small rotations, and ImageNet normalization), ensuring that differences in performance were attributable primarily to the backbone architecture rather than to differing data pipelines.
    \item Training was performed using binary cross-entropy loss and Adam optimizer. The~number of epochs, batch size, and learning-rate schedule were kept comparable to those used for the CvT models to provide a fair comparison.
    \item The train/validation/test splits were also the same as the corresponding CornViT models to ensure a fair comparison.
\end{itemize}

Results from Table~\ref{tab_baseline_accuracies} confirm that DenseNet-121 provides a stronger baseline than ResNet-50 across all three tasks, likely due to its enhanced feature reuse and gradient flow. Nevertheless, the proposed CvT-based CornViT models achieve higher accuracies in all stages (93.76\%, 94.11\%, and 91.12\%, respectively), highlighting the benefit of transformer-based global reasoning and the hierarchical design for fine-grained kernel classification.

\begin{table}[ht]
    \centering
    \caption{{Baseline} 
 accuracies of ResNet-50 and DenseNet-121 on the same test sets used for \mbox{CornViT evaluation}.}
    \label{tab_baseline_accuracies}
    \begin{tabularx}{\columnwidth}{Xcc}
        \toprule
        \textbf{Stage} & \textbf{ResNet-50} & \textbf{DenseNet-121} \\
        \midrule
        Stage 1            & 76.56\% & 86.56\% \\
        Stage 2            & 78.21\% & 87.05\% \\
        Stage 3            & 81.02\% & 89.38\% \\
        \bottomrule
    \end{tabularx}
\end{table}

\subsection{Stage 1---Pure vs. Impure Classification}

\begin{figure*}[ht]
    \includegraphics[width=6.5in]{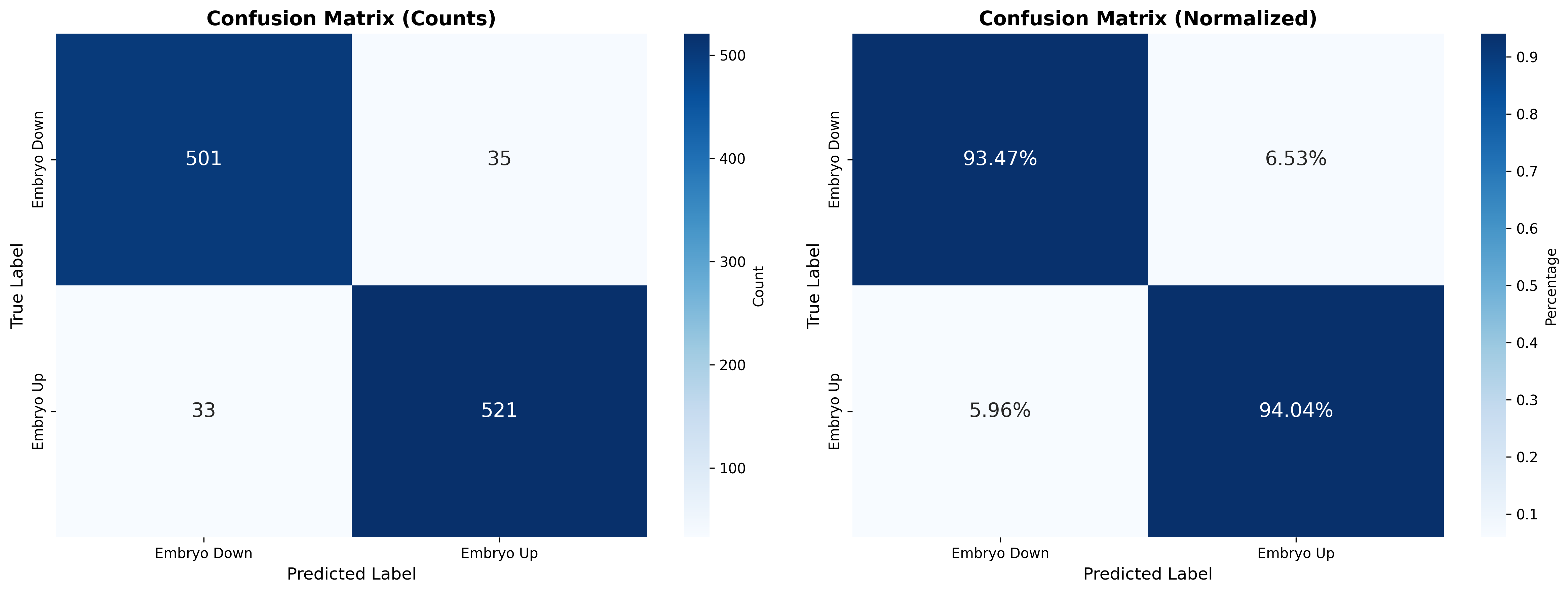}
    \caption{Confusion matrix for Stage~1, showing the distribution of true vs.\ predicted labels for the test set.}
    \label{fig_cm1}
\end{figure*}

Stage 1 distinguishes impure from pure kernels across 1090 test images. Table~\ref{tab_stage1_report} summarizes per-class metrics, and the corresponding confusion matrix is shown in Figure~\ref{fig_cm1}. Performance is well-balanced across classes, with both pure and impure kernels achieving F1 scores above 0.93. This is important in practice: false positives (impure labeled as pure) can contaminate subsequent grading, while false negatives (pure labeled as impure) reduce usable yield. The symmetry of Precision and Recall suggests that CornViT Stage 1 maintains a good trade-off between sensitivity and specificity.

\begin{table*}[ht]
    \centering
    \caption{Classification report for Stage~1 (Impure vs.\ Pure).}
    \label{tab_stage1_report}
    \begin{tabularx}{0.62\textwidth}{lcccc}
        \toprule
        \textbf{Class} & \textbf{Precision} & \textbf{Recall} & \textbf{F1-Score} & \textbf{Support} \\
        \midrule
        Pure           & 0.9370 & 0.9404 & 0.9387 & 554 \\
        Impure         & 0.9382 & 0.9347 & 0.9365 & 536 \\
        \midrule
        Accuracy 
        & --     & --     & 0.9376 & 1090 \\
        Macro avg      & 0.9376 & 0.9375 & 0.9376 & 1090 \\
        Weighted avg   & 0.9376 & 0.9376 & 0.9376 & 1090 \\
        \bottomrule
    \end{tabularx}
\end{table*}

Compared to the CNN baselines, the CvT-based model improves absolute accuracy by roughly 7--17 percentage points and substantially increases F1 scores, highlighting the benefit of transformer-based global reasoning even for relatively simple binary tasks.

\subsection{Stage 2---Shape Classification}

\begin{figure*}[ht]
    \includegraphics[width=6.5in]{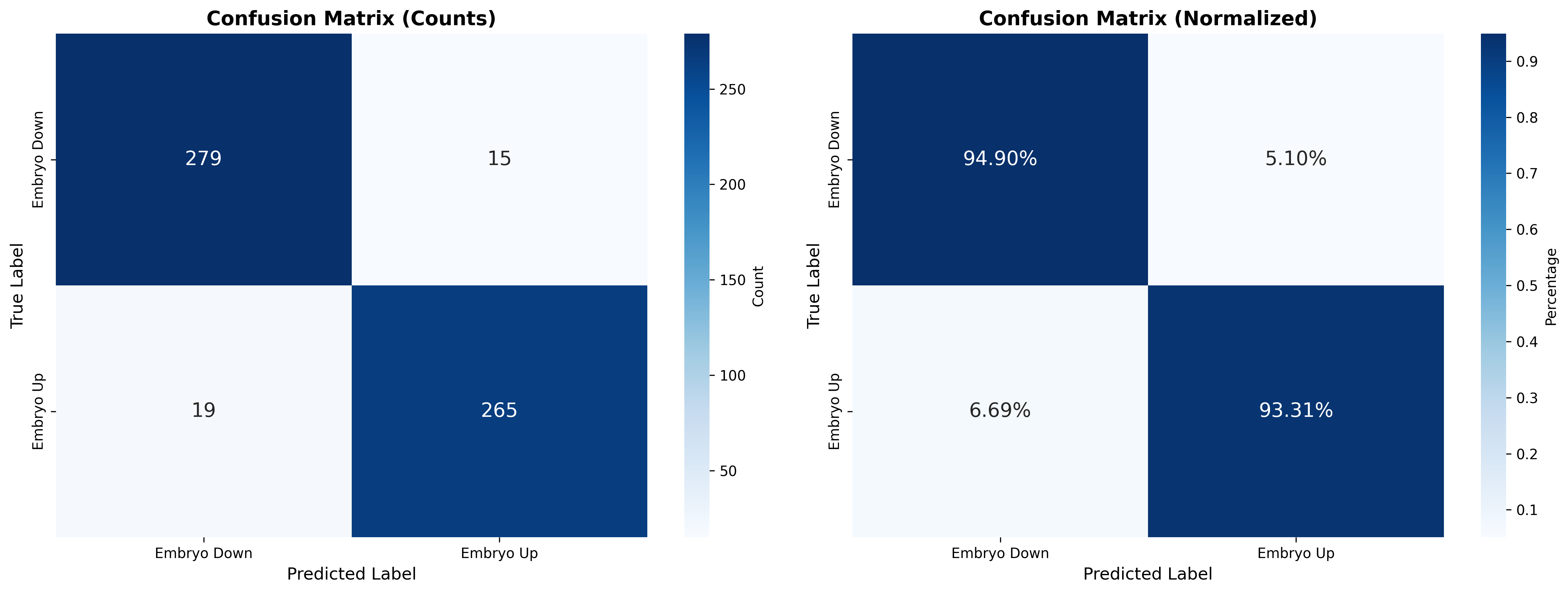}
    \caption{Confusion matrix for Stage~2, showing the distribution of true vs.\ predicted labels for the test set.}
    \label{fig_cm2}
\end{figure*}

Stage 2 receives pure kernels and predicts flat versus round morphology on a test set of 578 images. Table~\ref{tab_stage2_report} reports the quantitative results, and the corresponding confusion matrix is shown in Figure~\ref{fig_cm2}. Accuracy improves slightly relative to Stage 1, and both shape classes show nearly identical F1 scores ($\approx$0.94). Shape classification is inherently more subtle than gross impurity detection, as it primarily depends on global geometry and kernel silhouette. The~high performance of Stage 2 indicates that the CvT backbone effectively captures these morphological cues, supporting its suitability for shape-sensitive \mbox{grading tasks}.

This performance clearly surpasses the ResNet-50 and DenseNet-121 baselines, likely because those architectures are less effective at modeling the long-range spatial context required for accurate kernel shape analysis.

\begin{table*}[ht]
    \centering
    \caption{Classification report for Stage~2 (Flat vs.\ Round).}
    \label{tab_stage2_report}
    \begin{tabularx}{0.62\textwidth}{lcccc}
        \toprule
        \textbf{Class} & \textbf{Precision} & \textbf{Recall} & \textbf{F1 Score} & \textbf{Support} \\
        \midrule
        Flat           & 0.9362 & 0.9489 & 0.9425 & 294 \\
        Round          & 0.9464 & 0.9330 & 0.9396 & 284 \\
        \midrule
        \textbf{{Accuracy}}     & --     & --     & \textbf{{0.9411}} & \textbf{{578}} \\
        Macro avg      & 0.9413 & 0.9409 & 0.9410 & 578 \\
        Weighted avg   & 0.9413 & 0.9405 & 0.9413 & 578 \\
        \bottomrule
    \end{tabularx}
\end{table*}

\subsection{Stage 3---Embryo Orientation Classification}
\begin{figure*}[ht]
    \includegraphics[width=6.5in]{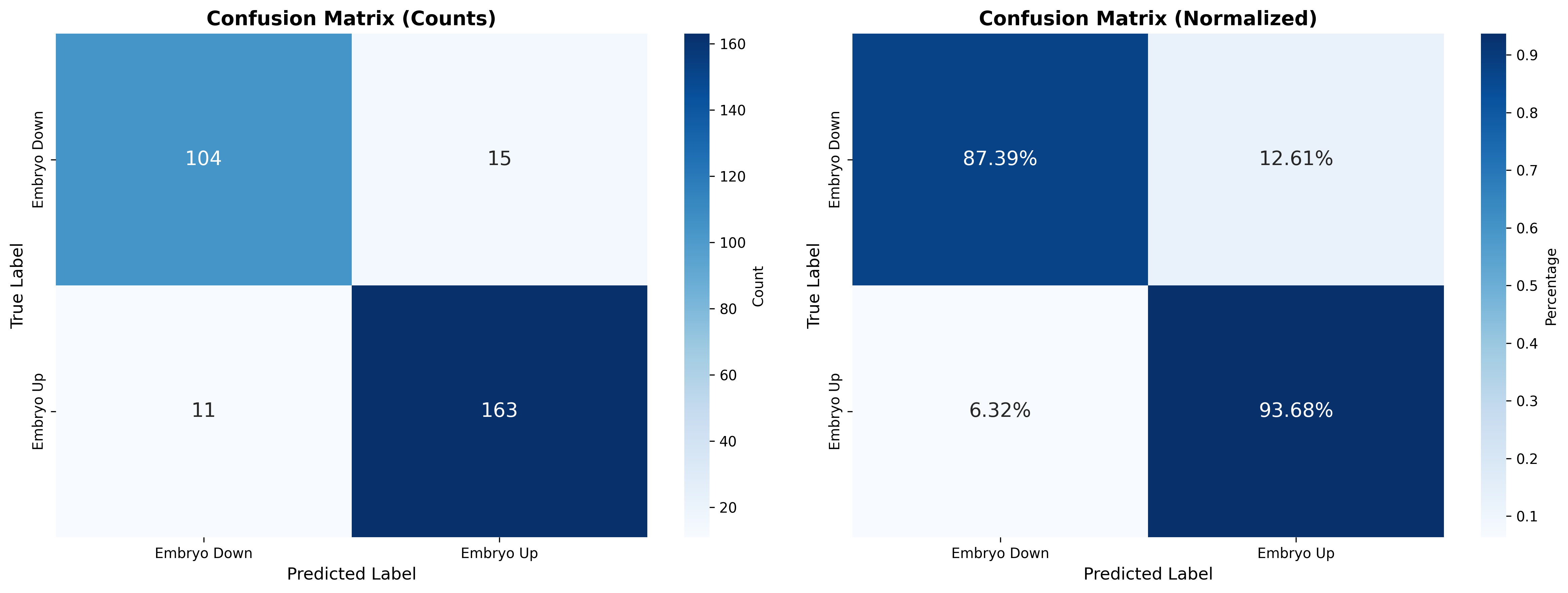}
    \caption{Confusion matrix for Stage~3, showing the distribution of true vs.\ predicted labels for the test set.}
    \label{fig_cm3}
\end{figure*}

Stage 3 is the most challenging step: Given pure, flat kernels, it predicts whether the embryo is facing up or down. The test set contains 293 images. Table~\ref{tab_stage3_report} shows per-class performance, and the corresponding confusion matrix is shown in Figure~\ref{fig_cm3}.

\begin{table*}[ht]
    \centering
    \caption{Classification report for Stage~3 (Embryo up vs.\ Embryo down).}
    \label{tab_stage3_report}
    \begin{tabularx}{0.62\textwidth}{lcccc}
    \toprule
    \textbf{Class} & \textbf{Precision} & \textbf{Recall} & \textbf{F1 Score} & \textbf{Support} \\
        \midrule
        Embryo down    & 0.9043 & 0.8739 & 0.8890 & 119 \\
        Embryo up      & 0.9157 & 0.9367 & 0.9261 & 174 \\
        \midrule
        \textbf{{Accuracy}}     & --     & --     & \textbf{{0.9112}} & \textbf{{293}} \\
        Macro avg      & 0.9100 & 0.9053 & 0.9076 & 293 \\
        Weighted avg   & 0.9110 & 0.9102 & 0.9100 & 293 \\
        \bottomrule
    \end{tabularx}
\end{table*}

Despite the fine-grained nature of the task and the smaller sample size, Stage 3 achieves over 91\% accuracy and macro F1 $\approx 0.91$. Embryo-up kernels are slightly easier to classify (higher Recall and F1), likely because embryo structures (e.g., scutellum, germ region) produce distinctive surface cues when facing the camera, whereas embryo-down kernels may resemble each other more subtly depending on lighting and angle.

The performance gap relative to the CNN baselines is even more pronounced here; CNNs find it difficult to distinguish such subtle orientation cues, whereas the CvT's attention mechanisms appear to better capture global texture and shape patterns that \mbox{signal orientation}.

{To summarize the comparative performance across all three stages, Table} \ref{tab_accuracy_comparisons} {reports test-set accuracies for CornViT and the two CNN baselines. CornViT achieves the highest accuracy in every stage, with gains of roughly 7 to 17 percentage points over ResNet-50 and about 2 to 7 percentage points over DenseNet-121.}

\begin{table*}[ht]
    \centering
    \caption{{Stage-wise comparison of CornViT and CNN baselines on the same test sets}}
    \label{tab_accuracy_comparisons}
    \begin{tabular}{lccc}
        \toprule
        \textbf{Stage/Task} & \textbf{CornViT Acc.} & \textbf{ResNet-50 Acc.} & \textbf{DenseNet-121 Acc.} \\
        \midrule
        Stage 1 (purity)              & \textbf{{93.76\%}} & 76.56\% & 86.56\% \\
        Stage 2 (shape)               & \textbf{{94.11\%}} & 78.21\% & 87.05\% \\
        Stage 3 (embryo orientation)  & \textbf{{91.12\%}} & 81.02\% & 89.38\% \\
        \bottomrule
    \end{tabular}
\end{table*}

\subsection{Overall Pipeline Behavior and Error Propagation}

Since the proposed framework, CornViT, is a hierarchical pipeline, errors in an early stage can propagate downstream. For~example, a kernel misclassified as impure in Stage 1 will never be considered for shape or orientation analysis. However, the very high Stage 1 accuracy ($\approx$93.8\%) limits the number of such cases.

One way to quantify end-to-end performance is to consider the effective accuracy for particularly important label combinations, such as ``pure, flat, embryo up.'' Assuming independence between stages, a rough lower bound on the probability of correctly classifying a kernel as pure, flat, and embryo up is
\begin{equation*}P(\text{correct all three}) \approx 0.9376 \times 0.9411 \times 0.9112 \approx 0.803.\end{equation*}

This indicates that 80\% of kernels that traverse all three stages could be classified correctly. In practice, the true end-to-end accuracy depends on the distribution of samples and correlation of errors across stages. {A more precise characterization of real-world performance would require a direct
end-to-end evaluation of the complete three-stage pipeline on a joint test set covering all purity–shape–orientation combinations, which we leave for future work.}

\subsection{Visual Analysis}

\begin{figure*}
    \includegraphics[width=6.5in]{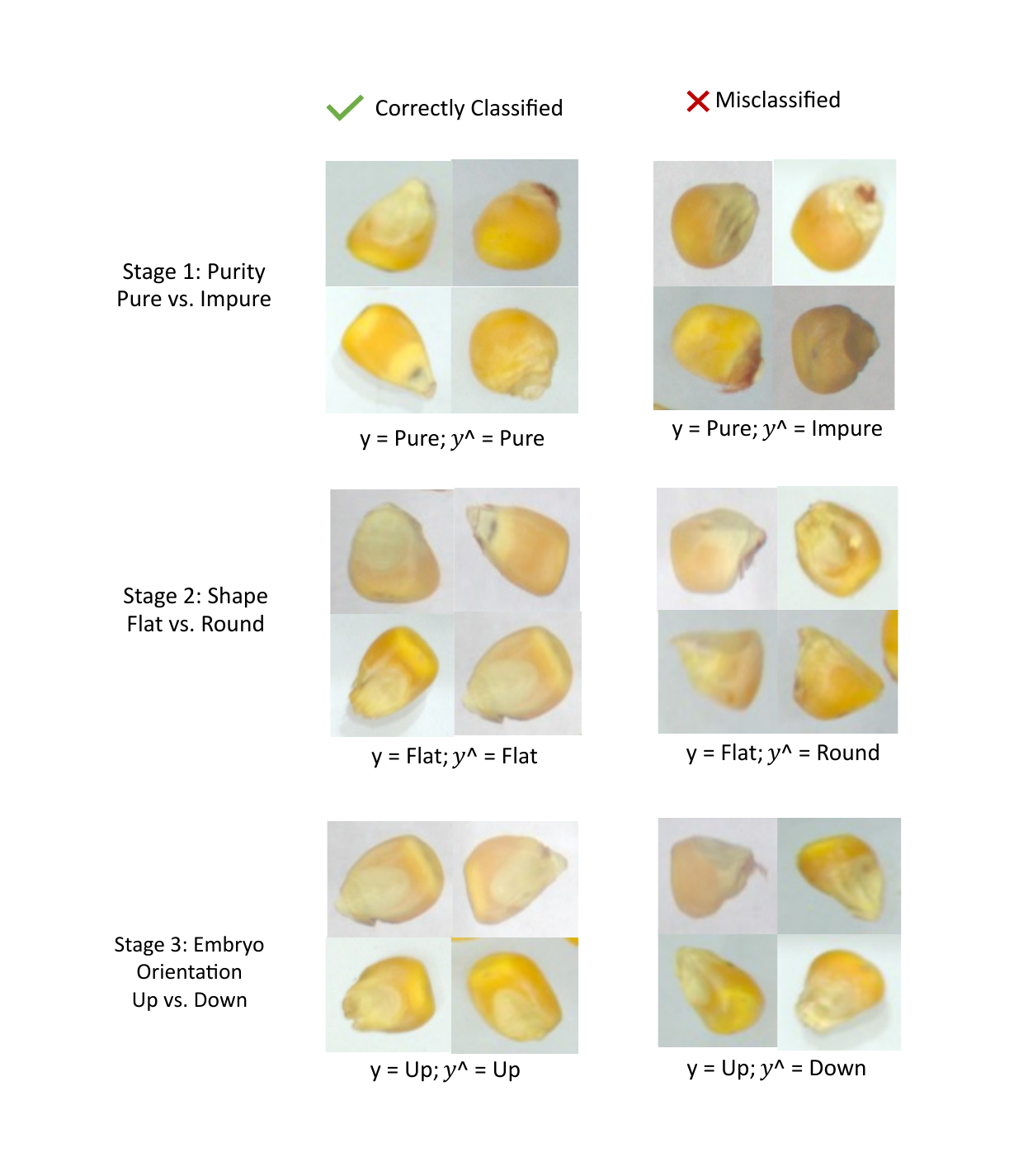}
    \caption{Representative CornViT predictions for each stage, including correctly classified and typical borderline misclassifications.}
    \label{fig_visual_analysis}
\end{figure*}

{To complement the quantitative metrics reported so far, we performed qualitative visual analyses of CornViT's predictions. Figure} \ref{fig_visual_analysis} {illustrates representative examples from each stage, including both correctly and incorrectly classified kernels where $y$ is the true label and $\hat{y}$ is the predicted label.

For Stage 1, typical misclassifications involve borderline impurities: kernels that carry small artifacts such as tip cap (red/brown patch, point where the kernel was attached to the cob) or have slight deviations from the idealized shape. These subtle irregularities can cause the model to hesitate or assign the wrong label. In Stage 2, errors are concentrated on kernels with intermediate shapes that lie between the ``flat'' and ``round'' prototypes. In Stage 3, the most challenging cases are kernels where the embryo is partially visible or where the embryo-facing side is only subtly different in appearance from the opposite side.}

\section{Web Application and System Integration}
\label{sec_web_app}

To make CornViT usable beyond the experimental setting, we implemented a ready-to-use web application that exposes the full hierarchical pipeline through a simple browser interface. This app is built with Python Flask (version 3.1.2) 
and bundles \cite{aslam2015efficient} all three trained CvT-13 models (purity, shape, embryo orientation) into a single end-to-end service.

\subsection{Implementation}
The system follows a standard client–server architecture. The~client is implemented using HTML \cite{pilgrim2010html5}, CSS \cite{meyer2006css}, JavaScript \cite{crockford2008javascript}, and provides a minimal interface that allows users to
\begin{itemize}
    \item Upload a single RGB image of a corn kernel;
    \item Invoke the classification process with a single button click;
    \item Inspect stage-wise predictions and confidence scores.
\end{itemize}

The server-side backend is implemented in Flask and is responsible for the following:
\begin{itemize}
    \item Loading the three CornViT models (Stage 1, 2, and 3) into memory at startup;
    \item Performing image validation and preprocessing;
    \item Orchestrating the hierarchical decision logic;
    \item Returning a structured JSON \cite{pezoa2016foundations} response to the client.
\end{itemize}

\subsection{User Workflow and Visualization}

\begin{figure}[!htbp]
    \centering
    \includegraphics[width=\columnwidth]{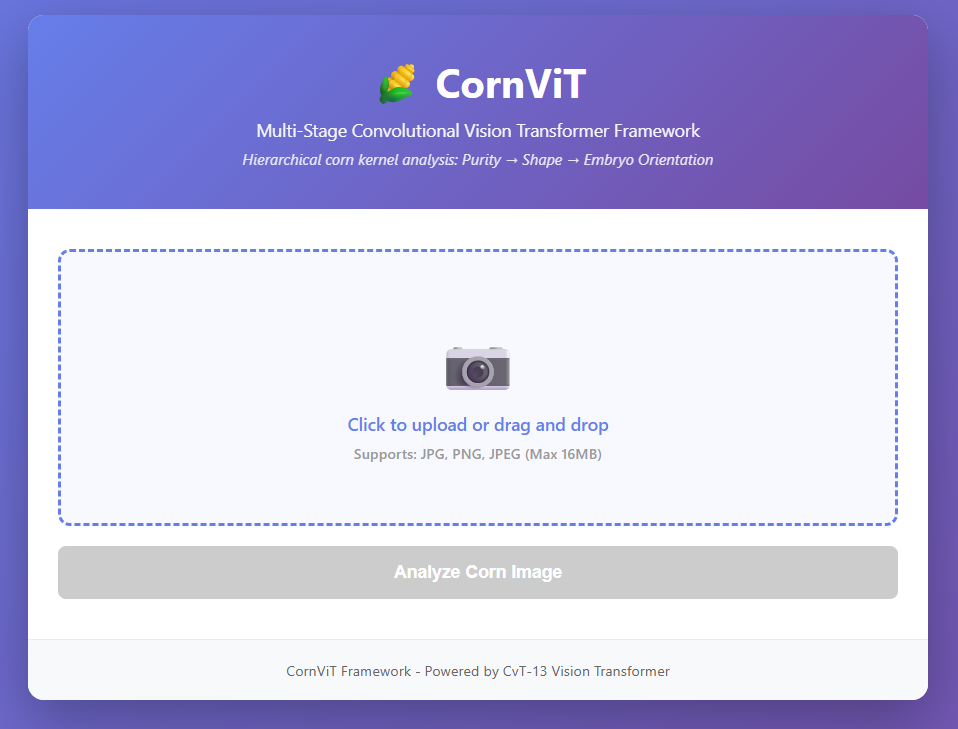}
    \caption{Landing page of the CornViT web interface.}
    \label{fig_landing_page}
\end{figure}

\begin{figure}[h]
    \centering
    \includegraphics[width=\columnwidth]{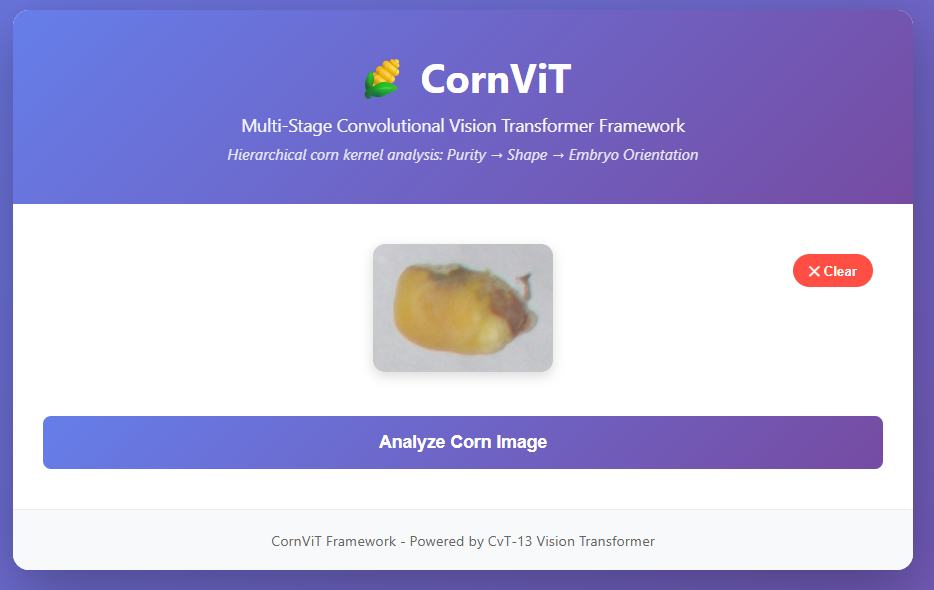}
    \caption{Screen showing the uploaded image and the analyze button.}
    \label{fig_upload_page}
\end{figure}

\begin{figure*}[ht]
    \centering
    \includegraphics[width=6in]{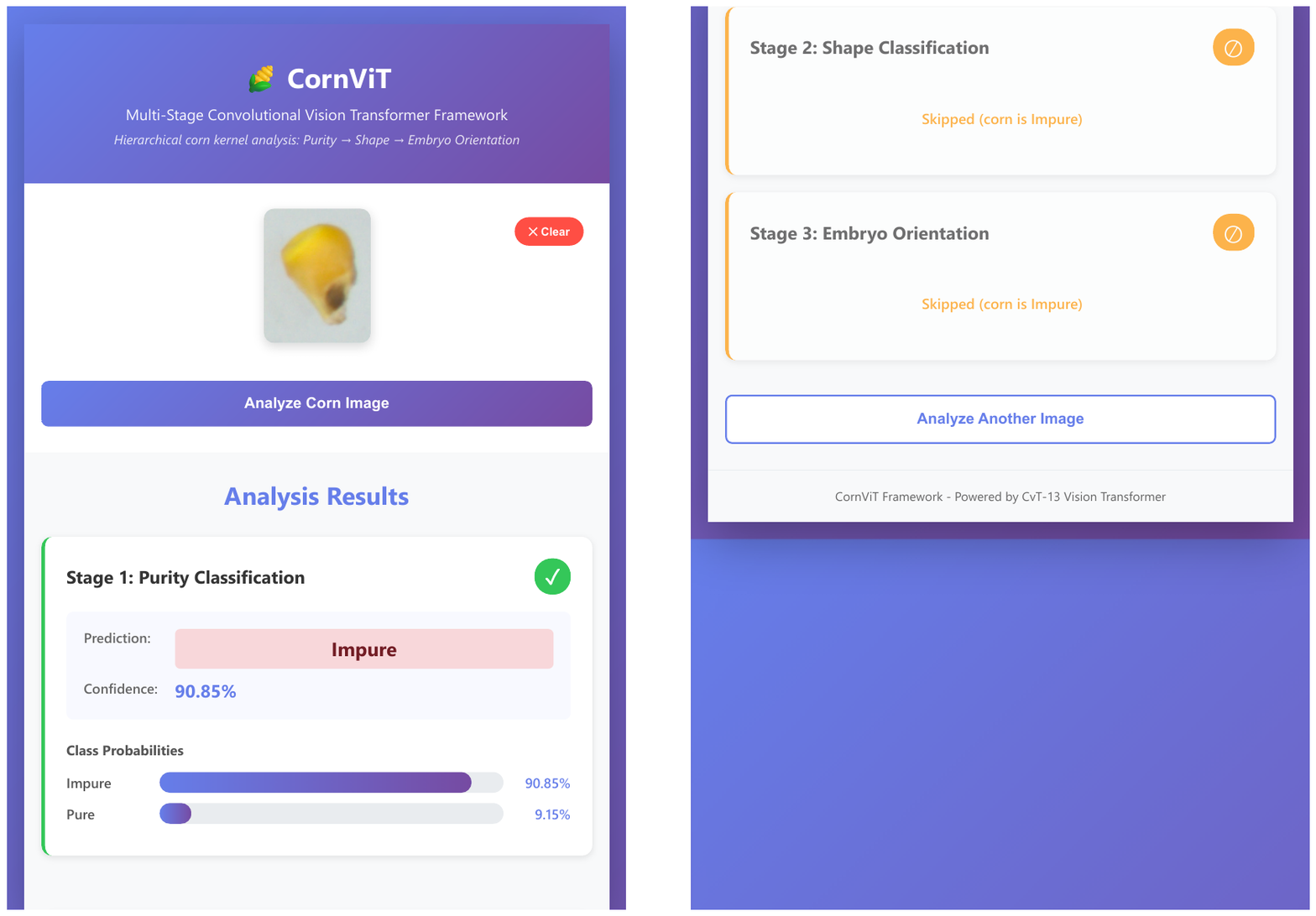}
    \caption{Example CornViT web-app result for an impure kernel. Stage 1 predicts Impure with high confidence, and the interface explicitly shows the corresponding class probabilities (Impure vs. Pure). Because the kernel is impure, Stages 2 (shape) and 3 (embryo orientation) are skipped and marked as not applicable.}
    \label{fig_impure_results}
\end{figure*}

\begin{figure*}[ht]
    \centering
    \includegraphics[width=6in]{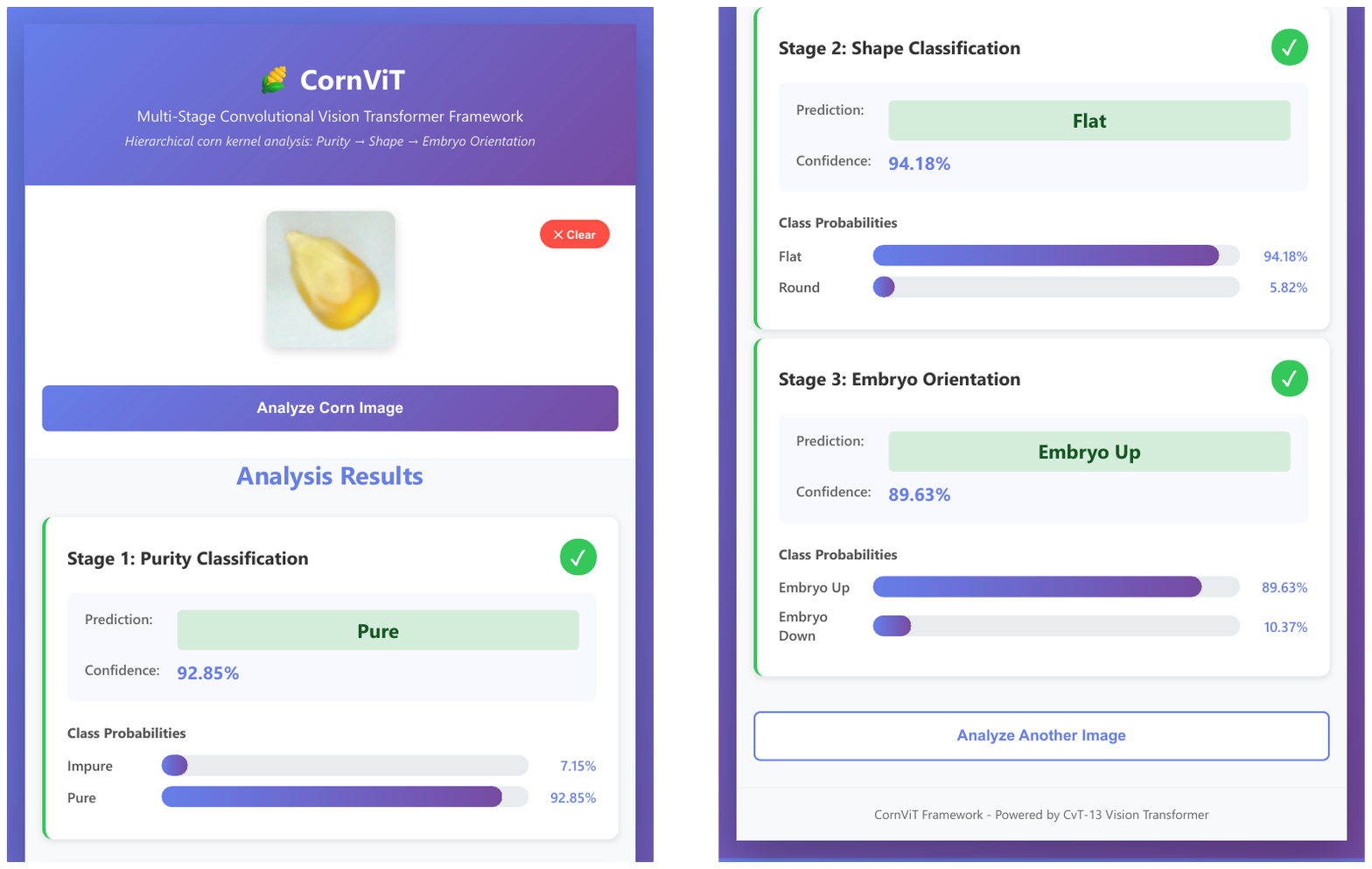}
    \caption{Example CornViT web-app result for a pure--flat kernel with the embryo facing up. Stage 1 predicts Pure with high confidence; Stage 2 then classifies the kernel as Flat and displays probabilities for the Flat and Round classes; Stage 3 finally predicts Embryo Up and provides probabilities for the Embryo Up and Embryo Down classes. All~three stages are executed.}
    \label{fig_pure_results}
\end{figure*}

From the user’s perspective, the workflow is illustrated as follows:
\begin{enumerate}
    \item Open the CornViT web application in a browser (landing page, Figure~\ref{fig_landing_page}).
    \item Upload or drag-and-drop an image of a single kernel (Figure~\ref{fig_upload_page}).
    \item Click the ``Analyze'' button.
    \item Inspect the stage-wise predictions and summary output returned by the system (Figures~\ref{fig_impure_results} and ~\ref{fig_pure_results}).
        \begin{itemize}
            \item Stage 1: Impure vs. Pure with confidence.
            \item Stage 2: Flat vs. Round (if pure) with confidence.
            \item Stage 3: Embryo up vs. Embryo down (if pure and flat) with confidence.
        \end{itemize}
\end{enumerate}

For each stage, the interface displays both the predicted class and its associated confidence, together with the full pair of class probabilities: Impure/Pure for Stage 1, Flat/Round for Stage 2, and Embryo Up/Embryo Down for Stage 3.

The interface is minimal, allowing agronomists, seed technicians, and industry users to operate the system without prior knowledge of deep learning or Python internals.


\subsection{Reproducibility and Deployment}

The complete source code for the web application is publicly available at 
 \url{https://github.com/SaiTeja-Erukude/CornViT-Client} (accessed on 19 December 2025). The~app is designed in a loosely coupled manner and is distributed together with

\begin{enumerate}
    \item The three trained CornViT model checkpoints (one for each stage);
    \item The CvT-13 configuration files;
    \item The preprocessing utilities and Flask routes.
\end{enumerate}

This packaging allows straightforward deployment on a workstation or server running a recent Python environment. The~same codebase can be containerized (e.g., via \mbox{Docker \cite{anderson2015docker}}), providing a bridge from the experimental CornViT models to practical, day-to-day kernel quality assessment.  Users may also substitute their own models while reusing the web application as a model-agnostic service.

\section{Discussion}
\label{sec_discussion}

The results demonstrate that the proposed CornViT framework can reliably reproduce human-style hierarchical reasoning for corn kernel grading. Across all three stages, the CvT-13 backbone paired with stage-specific binary heads achieved high and well-balanced performance, {with accuracies exceeding 91\%}. This~indicates that the combination of convolutional token embedding and convolutional self-attention projections provides a strong inductive bias for capturing both local texture and global morphology in kernel images.

A first observation is that the hierarchical design is both effective and practical. Rather than forcing the model to infer a composite label in a single step, CornViT decomposes grading into three interpretable decisions: whether the kernel is pure, what its gross shape is, and, for pure--flat kernels, how the embryo is oriented. Each decision aligns with a question human graders routinely ask, making the intermediate outputs meaningful in their own right. {The high and symmetric Precision--Recall values in Stages 1 and 2 suggest that these initial filters are robust, providing a stable foundation for more subtle orientation analysis in Stage 3.}

The comparison with CNN baselines further highlights the benefits of transformer-based backbones in this domain. {Under identical training conditions, CornViT consistently outperforms strong CNN architectures such as ResNet-50 and DenseNet-121 across all stages, with the largest gains observed in the embryo-orientation task, where discrimination depends on subtle cues in kernel surface structure and silhouette.} The results suggest that convolution-augmented self-attention is especially well-suited to tasks that combine fine-grained texture analysis with global shape reasoning.

The stage-wise dataset design also plays a crucial role. By constructing separate, carefully curated datasets for purity, shape, and embryo orientation, we avoid conflating label noise in the original source with model capacity. Each stage can be trained and evaluated on labels tailored to its specific decision, enabling a cleaner analysis of model behavior and error modes. At the same time, the three datasets can be reassembled to study error propagation in the full pipeline, e.g., how misclassifications in Stage 1 affect downstream shape and orientation predictions.

From a deployment perspective, the hierarchical structure and the accompanying Flask-based web application make the system attractive for practical seed quality workflows. The~web interface exposes stage-wise predictions and confidences through a simple browser front-end, lowering the barrier to adoption for agronomists and seed technicians who may not be familiar with deep learning frameworks. The~ability to skip later stages when a kernel is clearly impure also saves computation in potential \mbox{high-throughput settings}.

{Beyond corn kernels, CornViT fits into a broader trend toward automated, image-based classification of agricultural products. Similar hierarchical pipelines could be designed for fruit or vegetable grading (e.g., separating healthy from damaged fruit before finer defect categorization) or for multi-stage assessment of silage quality, where stages might reflect purity, particle-size characteristics, and kernel processing score. In this sense, CornViT complements recent work such as Succulent-YOLO} \cite{rs17132219}, {which uses a modern YOLO-based pipeline with CLIP-enhanced features for agricultural product classification from UAV imagery, and the 2022 study on real-time estimation of corn silage kernel processing score from image data} \cite{ROCHA2022107415}, {both of which demonstrate how deep learning can support decision-making across the crop production chain.}

At the same time, several limitations merit discussion. First, the imaging setup is relatively controlled, with a single general style of background, lighting, and camera placement. The~extent to which CornViT generalizes to other cameras, lighting conditions, or varieties remains to be fully explored. Second, the current model operates on single kernels placed on a uniform background; extending the approach to multi-kernel scenes or conveyor-belt imagery would require additional detection or segmentation modules. Third, the independence assumption used in our rough joint-accuracy estimate may not hold perfectly in practice, as errors across stages can be correlated. Finally, although head-only fine-tuning is computationally efficient and performed well here, it constrains the model's ability to adapt to the fine-grained visual differences that drive the Stage 3 embryo-orientation decision. More flexible strategies, such as selectively unfreezing the final CvT stage, applying layer-wise learning-rate decay, could enable the model to better capture subtle structural cues associated with embryo position. 

Future work will investigate several directions, including progressive unfreezing and low-rank adaptation (LoRA), as alternatives that balance stability with representational flexibility. They will also examine joint multi-task training with a shared backbone and multiple heads, which may improve parameter efficiency and exploit shared structure across stages. Domain adaptation and data augmentation strategies could be explored to improve robustness to new imaging setups and additional corn varieties. Incorporating attention or saliency visualizations directly into the web interface may further enhance interpretability for end users. Finally, extending the hierarchical framework to other crop species and quality attributes such as damage, disease, or varietal classification could broaden its relevance to precision agriculture and seed-processing pipelines.

\section{Conclusions}
\label{sec_conclusions}

This paper presented CornViT, a multi-stage Convolutional Vision Transformer framework for hierarchical corn kernel analysis. By explicitly mirroring the reasoning process of human seed analysts through three sequential decisions: purity (impure vs. pure), shape (flat vs. round), and embryo orientation (up vs. down), CornViT delivers accurate and interpretable kernel-level grading from RGB images. Across dedicated test sets, the three CvT-13 models achieved stage-wise accuracies of 93.76\%, 94.11\%, and 91.12\%, respectively, with strong per-class F1 scores, including for the more challenging embryo-orientation task.

Compared with strong CNN baselines, CornViT consistently delivers higher stage-wise performance, underscoring the benefits of convolution-augmented self-attention for capturing global morphology and subtle surface cues within a hierarchical \mbox{decision framework}.

A further contribution of this work is the construction of a ready-to-use, stage-wise annotated dataset tailored to hierarchical kernel classification. Each image is labeled for purity, morphology, and, where applicable, embryo orientation, enabling both isolated stage-wise training and end-to-end pipeline evaluation. Together with the curated datasets, the lightweight Flask-based web application that encapsulates the full CornViT pipeline makes the approach readily deployable in laboratory and industrial environments.

Future research directions include investigating {progressive unfreezing,} joint multi-task training of a shared backbone, exploring more advanced fine-tuning strategies, and improving robustness to diverse imaging conditions and corn varieties. Extending the framework to multi-kernel scenes, conveyor-based inspection, and other crop species would further enhance its impact on smart seed processing and precision agriculture. Overall, the CornViT framework and accompanying dataset provide a solid foundation for accurate, interpretable, and deployable seed- and kernel-level quality control systems.

\bibliographystyle{apalike}
\bibliography{main}

\end{document}